# Ensembled Correlation Between Liver Analysis Outputs

S. E. Seker, Y. Unal, Z. Erdem, and H. Erdinc Kocer

*Abstract*—Data mining techniques on the biological analysis are spreading for most of the areas including the health care and medical information. We have applied the data mining techniques, such as KNN, SVM, MLP or decision trees over a unique dataset, which is collected from 16,380 analysis results for a year. Furthermore we have also used meta-classifiers to question the increased correlation rate between the liver disorder and the liver analysis outputs. The results show that there is a correlation among ALT, AST, Billirubin Direct and Billirubin Total down to 15% of error rate. Also the correlation coefficient is up to 94%. This makes possible to predict the analysis results from each other or disease patterns can be applied over the linear correlation of the parameters.

## I. Introduction

MEDICAL analysis has shown with using machine learning techniques for two decades. Learning of creation machine is beneficial on medical analysis because it provides to decrease human resources and their cost, and increase the accuracy of diagnosis [1].

The liver is an effective organ at neutralizing and expelling toxins from the body. If the amount of toxins exceeds the organs functioning capacity, cells of affected areas in the organ will experience cell damage. Some emerging substances and enzymes will be released into the bloodstream. While the patient is being diagnose, enzymes levels in the blood will be analyzed. Both elevated enzyme levels and the varied effects of different alcohol levels on different patients can result in inaccurate diagnosis [1],[2].

This study is built on multiple methodologies studied on the analysis related to the liver health. The methodologies are well-known classification and clustering algorithms to find out the correlation between the four different liver functioning tests, ALT, AST, Bilirubin Direct (BD) and Bilirubin Total(BT). The values are suitable for the correaltion and this correlation can be useful to autmoatically identify disease patterns or temporal monitoring of the liver disease status.

Liver function tests are operated to give information about the patient's liver. The tests can be applied for several reasons.

- Screening to identify the liver dsyfunction
- Pattern of disease to recognize the type of disease
- Asses severity to understand how sever is the disease
- Follow up to keep track of the liver disease [2].

In this study we propose a correlation among the major test parameters using ensemble classification algorithm over three classical classification algorithms which are already proven the success on liver disorder.

## II. Related Works

If we look at the literature regarding this subject, almost all of the studies mentioned below have been carried out based on the Liver Disorders Data Set at the UCI database. Data mining algorithms were implemented to the data as part of these studies and in the other part of the study, other machine methods of learning were used.

Bendi Venkata Ramana et. al., used two different database using naive bayes, C4.5, back probagation, Neural network and SVM that are the classification method in their studies [3].

Studies conducted by P.Rajeswari, G.Sophia Reena based on the dataset from UCI were classified by the WEKA program through the use of Naive bayes, Kstar ve FT tree algorithms [4].

In the study conducted by Hyontai sug [5] used 3 different data mining algorithms (decision tree, C4.5 and CART) in the dataset obtained from the UCI and with the method of oversampling, he attempted to increase prediction accuracy.

In the study conducted by Rong-Ho Lin used classification and regression tree (CART) and case-based reasoning (CBR) techniques [6] which were obtained from medical center in Taiwan from 2005 to 2006.

Kim et. al., adopted a prediction model including logistic regression, a decision tree and a neural network to analyze the risk factors of liver disease [7].

Newton Cheung, using C4.5 %65.59 [1], Newton Cheung, using Naive Bayes %63.39 [1], Newton Cheung, BNND (Bayesian Network with Naive Dependence) %61.83 [1], Newton Cheung, BNNF (Bayesian Network with Naive Dependence & Feature Selection) using %61.42 [1], Tony Van

S. E. Seker is with the Computer Engineering Dept. of Istanbul University, Istanbul , TURKEY (corresponding author to provide phone: +90-532-446-7882; e-mail: academic@sadievrenseker.com).
Y. Unal, is with the Computer and Inst. Tech. Dept. of Amasya University, Amasya, TURKEY (e-mail: yavuz.unal@amasya.edu.tr).
Z. Erdem Turkish National Science Foundation, Ankara, TURKEY, (e-mail: zeki.erdem@tubitak.gov.tr).
H. E. Kocer is with the Electronic and Computer Edu. Dept. of Selcuk University, Konya, TURKEY(ekocer@selcuk.edu.tr).

Gestel et. al., VMC (Vector Machine Classifiers) using %69.7 [8], Yuh-Jye Lee ve O.L. Mangasarian, SSVM (Smooth Support Vector Machine) using %70.33 [9], Yuh-Jye Lee ve O.L. Mangasarian, RSVM using %74.86 [10], D.T. Pham et. Al., Inductive Learning using %53.76 [11], M. Yalçın ve T. Yıldırım, MLP using %74.56 [2], M. Yalçın ve T. Yıldırım, PNN using %73,.86 [2], M. Yalçın ve T. Yıldırım, GRNN using %60.68 [2], M. Yalçın and T. Yıldırım, RBF using %58.99 [2], K. Polat and his friends AIRS (Supervised Artificial Immune System)using %81[12], again K. Polat and his friends FW-AIRS using %83.38[13], M. Neshat and A. E. Zadeh, hopfield neural network and fuzzy hopfield neural netwok using have been achieved 91% success rate[14].

## III. BACKGROUND

The background section of this study can be divided into two categories. The first category holds the background of the data mining techniques and the second is the background information about the laboratory tests.

Decision Stump, M5P, REPTree, Bagging, MLP, SVM and KNN data mining techniques used in this study.

After the model tree proposition [15] by Quinlann in 1992, which is capable of classification over the continous data sets via class functions instead of discrete class labels, there has been several improvements on those trees.

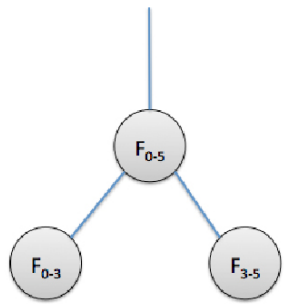

Fig. 1. M5 tree structure

M5 is a variation of model trees where the selection method of attributes is replaced with predictive attribute instead of theoretic metrics in classical model trees [16].
Later on the model tree concept is utilized for extracting the rules from the data set [17].

M5P, tries to generate a model tree, where each of the leaf nodes keeps a function over the continuous data set with the maximum coverage.

For example a data set equally distributed between 0 and 5 can be handled with a model tree demonstrated on Figure 1 if the level of tree is limited with 2.

REPTree is a regression tree built on the continuous data set. Different than the model tree, regression trees hold the data (instances) in each leaf and the nodes hold the rules to sperate the values.[18] The tree building is done over the information gain and variance reduction. Also REPTree utilizes the prunning and reduced-error prunning for the performance issues [19].

Decision stump [20], as a decision tree is one of the most simplified version of the decision tree learning algorithms. The level of the tree is limited with 1 only and most crucial rule is kept on the only internal node to divide the dataset into sub-sets.

Bagging [21] is the short cut of boostrap aggregation method and it can be considered as a meta algorithm working as an ensemble function on the machine learning problems. Bagging can be considered as a voting algorithm in small pieces of the data set. Algorithm starts with dividing the data set into small samplings where each sampling is smaller than the dataset itself and for each sample the sampling is done uniformly with replacement. After creating the sample groups from the data set, bagging applies data mining techniques and and monitors if any observation is repeating in the rest of the sampling groups. The most repeated learning outcome is considered as more crucial learning outcome like in the majority vote learning.

Relevance Vector Machines (RVM) is a special case of GP and the methodology of RVM is close to the support vector machine (SVM) [22],[23].

Purpose of ANN studies is adapting the biological neural networks into data processing. Multi-layer perceptron (MLP) is a developed version of ANN which organizes the neurons into layers as input, output or hidden layers [24],[25].

The last classifier implemented in this study is KNN (k-nearest neighborhood). The k, c-neighborhood (or k, c(x) in short) of an U-outlier $x$ is the set of $k$ class $c$ instances that are nearest to $x$ (k-nearest class c neighbors of x).

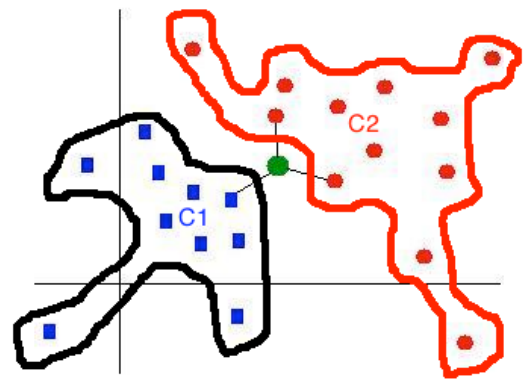

Fig. 2. KNN grouping

The K-NN [14] is explained in Figure 2. Here k is a user defined parameter. For example, k , c1(x) of an U-outliers x is the k-nearest class c1 neighbors of x.

Let $\overline{D}_{Cout,q}(x)$ be the mean distance of a U-outlier x to its k-nearest U-outlier neighbors. Also, let $\overline{D}_{C,q}(x)$ be the mean distance from $x$ to its $k,c(x)$, and let $\overline{D}_{C\min,q}(x)$ be the minimum among all $\overline{D}_C, q(x)$, $c \in$ {Set of existing classes}. In order words, $k$, $c_{\min}$ is the nearest existing class neighborhood of $x$. Then k-NSC of $x$ is given in equation (1).

$$k - NSC(x) = \frac{\overline{D}_{C\min,q}(x) - \overline{D}_{Cout,q}(x)}{\max(\overline{D}_{C\min,q}(x), \overline{D}_{Cout,q}(x))} \quad (1)$$

### A. Ensemble Classification

We have implemented MaVL (Majority Vote Learning) [24] based ensemble method to combine three different classification methods. MV can be considered as a meta classifier which works over the classifiers like KNN, C4.5 or SVM in our case.

Let $S_i \in S$ where S is the set of classifiers and let $C_i \in C$ where C is the set of classes,

$$C(x) = argmax_i \sum_{j=1}^{B} w_j I\left(S_j(x) = i\right) \quad (2)$$

Where $w_j$ is the weight of each indicator function I(•) which is added into the equation for normalization and the weights of each classifier is equal in our model.

MaVL, gets the summation for each of the classifier's vote and the sample is classified into the class with the highest vote.

### B. Liver Function Tests

Commonly used tests are available to see how well the liver functions. Some of these test include:

ALT : Alaine transaminase (ALT) is an enzyme and is found in the liver at high levels. It is an indicator of whether or not there is liver damage. If the blood at high level, it indicates liver damage. It is measured by blood test. The normal range is 10 to 40 international units per liter (IU/L) [25].

AST: AST (aspartat aminotransferaz) is an enzyme and is in liver, heart and muscle cell at high levels. At the same time, it is smaller at the other tissues. It is measured by blood test. The test is done with other tests (for example; ALP and bilirubin) to liver disease diagnosis and monitoring. The normal range is 10 to 34 IU/L.( IU/L = international units per liter) [27],[28].

Bilirubin: Bilirubin is in the bile and a yellowish fluid generated by the liver. A small amount of old red blood cells is replaced with new blood cells for each day. This old blood cellsis removed and than bilirubin is replaced. With help of bilirubin the liver helps to rid the body of destroyed old red blood cells though the stool. It is measured by blood test. High amount of bilirubin in the blood may cause jaundice. The normal level in the blood is as follows: Direct (also called conjugated) bilirubin: 0 to 0.3 mg/dL. Total bilirubin: 0.3 to 1.9 mg/dL. (mg/dL = milligrams per deciliter)[27],[28].

## IV. MATERIAL AND METHODS

*Dataset*

Properties of the data set is given in Table I.

TABLE I.    MAJOR TEST PARAMETER PROPERTIES

| Attributes | Min. | Max. | Mean | StdDev |
|---|---|---|---|---|
| ALT | 6 | 2.708 | 23.475 | *45.203* |
| AST | 3 | 4.202 | 23.729 | *65.335* |
| Bilirubin Direct | 0.01 | 46.388 | 68.195 | *1553.409* |
| Bilirubin Total | 0.01 | 44.593 | 72.887 | *1,576.013* |

The test results are collected from a state hospital during a 1 year long study

## V. EVALUATION

In this section, the evaluation of data mining techniques on the data set are evaluated via relative absolute error (RAE) and root relative square error (RRSE) calculations. We also provide a correlation coefficient, which is calculated via Pearson product-moment correlation coefficient method. (PPMCC) The results are provided just after giving the details of the evaluation calculations.

### A. Relative Absolut Error (RAE) Equations

The third error calculation method is RAE (Relative Absolute Error) and the calculation is given in equation (3) [29].

$$E_i = \frac{\sum_{j=1}^{n} |P_{(ij)} - T_j|}{\sum_{j=1}^{n} |T_j - \overline{T}|} \quad (3)$$

$P_{(ij)}$ is the value predicted by the individual program $i$ for sample case $j$ (out of n sample cases), $T_j$ is the target value for sample case $j$, and $\overline{T}$ is given by the equation (4) [29]:

$$\overline{T} = \frac{1}{n}\sum_{j=1}^{n} T_j \quad (4)$$

For a perfect fit, the numerator is equal to 0 and $E_i = 0$ So, the $E_i$ index ranges from 0 to infinity, with 0 corresponding to the ideal.

### B. Root Relative Square Error (RRSE)

Also the results are interpreted by using a second error calculation method RRSE (Root Relative Squared Error) and the calculation is given in equation (5) [30].

$$x_{rrse} = \sqrt{\frac{\sum_{j=1}^{n}(P_{ij} - T_j)}{\sum_{j=1}^{n}(T_j - \overline{T})^2}} \quad (5)$$

Where $P_{(ij)}$ is the value predicted for the sample case $j$, $T_j$ is the target value for sample case $j$ and $\overline{T}$ is calculated by equation (6) [30].

$$\overline{T} = \frac{1}{n}\sum_{j=1}^{n} T_J \quad (6)$$

The RRSE value ranges from 0 to $\infty$, with 0 corresponding to ideal.

*C. Pearson Product-Moment Correlation Coefficient Method (PPMCC)*

Pearson prodcut-momment correlation coefficient is a real number between +1 and -1 to measure the linear correlation between two variables. [30].

In most simple form, it is symbolized with (ρ) and can be considered as the covariance of both variables divided by the standard deviations of the variables.

$$P_{X,Y} = \frac{\text{cov}(X,Y)}{\sigma_x \sigma_y} \quad (7)$$

If the covariance is rewritten as expected value:

$$P_{X,Y} = \frac{E[(X - \mu_x)(Y - \mu_y)]}{\sigma_x \sigma_y} \quad (8)$$

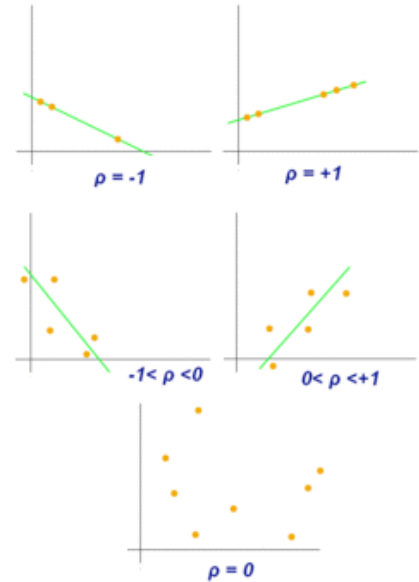

Fig. 3. examples of scatter diagrams with different values of correlation coefficient

The value of 1 means the samples are exactly on the same line and the slope of line is positive. The value of -1 means the samples are on the same line but the slope is negative this time. Any value between 0 an +1 indicates the slope is positive and the points are not exactly on the same line but a linearity can be claimed with a line passing close to each of the samples. On the other hand if drawing a line is impossible among the samples than the ρ value is 0.

*D. Evaluation of Data Mining Techniques*

During this study, 9 different data mining technique has been applied over the dataset. The methods and the error rates are given in Table II.

TABLE II.    PERFORMANCE STUDY OF ALGORITHM

|  | PPMCC | RAE (%) | RRSE(%) |
|---|---|---|---|
| KNN, N=1 | 0.8855 | 20.48 | 48.68 |
| KNN, N=3 | 0.9102 | 17.83 | 41.43 |
| SVM | 0.8692 | 20.78 | 50.21 |
| Desicion Stump | 0.9390 | 15.43 | 34.38 |
| M5P | 0.9243 | 19.31 | 38.42 |
| REPTree | 0.9340 | 15.84 | 35.72 |
| MLP | 0.9305 | 18.58 | 36.62 |
| Simple Linear Regression | 0.8686 | 29.37 | 49.55 |
| Bagging | 0.9303 | 16.79 | 36.68 |
| **MaVL** | **0.9423** | **16.66** | **35.22** |

## VI. Conclusion

We believe this study can be useful for further studies like reducing the number of analysis, since the prediction can be correlated and furthermore the correlation can be utilized for detecting the anomaly on the analysis.